# Efficient approach of using CNN based pretrained model in Bangla handwritten digit recognition

*Islam M.[1], Shuvo SA.[2], Nipun MS.[3], Bin Sulaiman R.[4]
Shaikh MM.[5], Nayeem J.[6], Haque Z.[7], Sourav MSU.[8]

[1]Southeast University, Dhaka,Bangladesh
[1]muntarinislam424@gmail.com

[2]Universität Rostock, Rostock, Germany
[2]shuvo.shabbirahmed@gmail.com

[3,4]University of Bedfordshire, Luton, United Kingdom
[3]musarrat.nipun@gmail.com
[4]rejwan.binsulaiman@gmail.com

[5]University of Portsmouth,Portsmouth , UK
[5]mostakshaikhcse@gmail.com

[6]University of Greenwich,London,United Kingdom
[6]nayeem3004@gmail.com

[7]Brandenburg University of Technology, Cottbus, Germany
[7]zubaer.haque@gmail.com

[8]Shandong University of Finance and Economics, Jinan, China.
[8]sakibsourav@outlook.com

**Abstract.** Due to digitalization in everyday life, the need for automatically recognizing handwritten digits is increasing. Handwritten digit recognition is essential for countless applications in various industries. Bengali ranks the fifth largest dialect in the world with 265 million speakers (Native and non-native combined) occupying 4 percent of the world population. Due to the complexity of Bengali writing in term of variety in shape, size, and writing style, researchers did not get better accuracy using supervised machine learning algorithms to date. Moreover, fewer studies have been done on Bangla handwritten digit recognition (BHwDR). In this paper, a novel CNN based pre-trained handwritten digit recognition model has been proposed which includes Resnet-50, Inception-v3, and EfficientNetB0 on NumtaDB dataset of 17 thousand instances with 10 classes. The Result outperformed the performance of other models to date with 97% accuracy in the 10-digit classes. Furthermore, we have evaluated the result of our model with other research study while suggesting future study.

## 1 Introduction

Bangla Handwritten digit recognition (BHwDR) is a domain of interest to the wide range of researchers to boost the effort of digitalization and add value to the 5th largest language in the world with around 4% of the world's total population [1]. The application of BHwDR system is wide from postal code digit recognition to license plate recognition, digit recognition in cheques in the banking system to exam paper registration no recognition..
Bangla handwritten digits and characters feature a more intricate arrangement of curvatures than characters in other languages like English or German. However, there are ten digits number system same as English or any other prominent languages in the world. but. Handwritten digit identification can be of great assistance to both personal and professional situations. In addition to this, it simplifies our lives by enabling us to tackle difficult challenges in a shorter amount of time.
The structure and appearance of the digits also influence the distinctiveness and variation of individual's handwriting. However, this is a challenging problem because there are a lot of variations in handwriting from person to person [2]. Therefore, we have come up with a distinct method for digit recognition, which will contribute to the development of the Bangla handwriting recognition sector. Our proposed model recognizes if an image is understandable or blurred from the handwritten image regardless of writing variation and appearance.

This study also compares various pre-trained models in terms of predicting Bangla handwritten recognition. The main contribution of this study is as follows:

- Build a deep learning model to recognize handwritten images of the digits of the Bengali Numerals.
- Effective CNN based pre-trained approach of detecting Bangali Handwritten digits.

Our proposed model will make the official work faster than ever. For instance, The government forms student information on exam papers are mostly in Bangla, including roll number, registration number, and postal code, which are often handwritten. These data are time-consuming and challenging to add to a digital platform, but handwritten digit recognition can be a good and easy way to digitize them. Security and police often try to locate specific vehicles based on their number plates; In automated parking lots, ticket generation is done by reading the number plate. A better solution can address this issue efficiently.
Given the variety of writing styles and shapes used in Bangla, handwritten digit recognition is one of the most difficult computer vision challenges. The one thing common with other languages is that Bangla Digit comprises ten main characters. However, the following figure 1 illustrates the variety of human ways of writing Bangla digits, demonstrating their rich diversity.

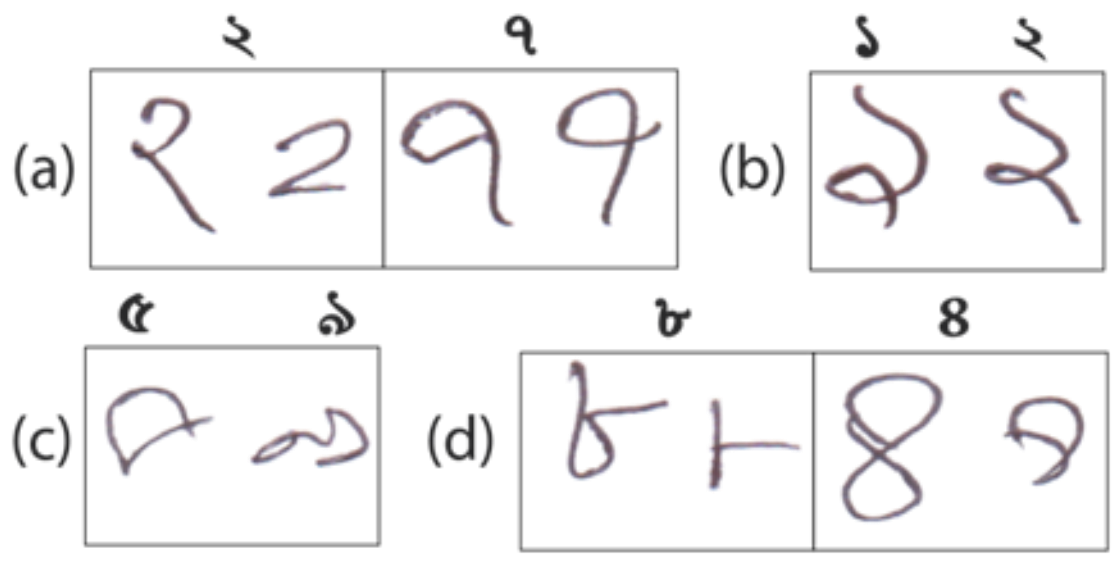

**Figure 1**: Diagram showing some of the challenges of Bangla digits.

Some demanding structural characteristics of the characters of the Bangla handwritten digits are described below:

1. Every writer has a unique writing style. It is very much distinct from each other. In Figure 1a there is a visual representation of how the same character can be written in diverse shapes and sizes.
2. Despite the fact that the characters come in a variety of sizes and shapes, some of them are introduced as puzzle characters because of their striking similarity in shape. Figure 1b shows a typical pair of character pairings from a problem..
3. The digit size that the writer creates plays an essential role and makes the digit lines crowded. Therefore, the digits look different than regular shapes. In Figure 1c, the digit size is small, making the line congested.
4. Digits containing circles in shape also cause difficulties. While writing, sometimes the circle turns a curly line. Figure 1d shows the rounded shapes depending into straight and curved lines.

To tackle this challenge the first thing would be to classify the digits as ০ and ১, ১ and ৭,২ and ১,৫ and ৬, etc.

The next parts of the paper are laid out as follows: Section 2 contains literature on recognizing and classifying different images in Bengali handwritten digits. Section 3 briefly summarizes the data set we collected. Section 4 describes our methodology in detail. The results of our experiment are presented in Section 5. Finally, in Section 6, we conclude our paper.

## 2 Literature Review

Numerous investigations have been made into the recognition of handwritten characters in certain widely used languages, including English [2], Chinese [3], Spanish [4], Arabic [5], Hindi [6], Farsi [7], Greek [8] and other languages. Bengali handwritten characters have, however, not been studied much. A minimal number of contributions have also been made in recognition of Bengali compound characters. Below we discuss various existing approaches in BHwDR using different methods.

### 2.1 Multi-layer Perceptron (MLP) Classifier

While working on MLP, the authors [9] discovered certain drawbacks, such as the loss function of MLP with hidden layers being non-convex due to the presence of many local minimums. Whereas another work [10] deployed different random weight initializations which leads to additional validation accuracy. In MLP, a number of hyperparameters, including the quantity of hidden neurons, layers, and iterations, can be modified. Complex handwritten numbers in Bangla are challenging to read because MLP is sensitive to feature scaling.

### 2.2 Inceptionv3

Inception-v3 based transfer learning has been used in the study [11] to highlight offline handwritten recognition for the Tamil language. This work demonstrates that the pre-trained model can surpass the earlier deep learning designs by achieving the recognition accuracy of 93.1 percent with the proper arrangement of transfer learning approach with Inception-v3. What is more, the work from Basri Rabeya et al [18] used AlexNet, Inceptionv3, MobileNet and CapsuleNet and reported their results where AlexNet demonstrates the best results on the highly augmented NumtaDB dataset. As Inception-v3 based CNN model developed by Tallapragada et al. [19] to recognize Greek handwritten characters depicts with 99% -accuracy, we evaluate this model with different parameter settings considering the effectiveness of this pre-trained CNN model.

### 2.3 Resnet50

Recent studies [12, 13] proposed using ResNet 50, a cutting-edge deep convolutional neural network model pretrained on the ImageNet dataset, the researchers employ various strategies yet unable to get desirable accuracy. The work [12] obtained an accuracy of 97.12% on the test set while having 47 epochs. Yet getting a decent accuracy, the average time for prediction is pretty higher. In the paper [13] authors identified scarcity of dataset has a huge impact in recognizing Bangla handwritten characters. However, these approaches show that restnet50 is potentially a better model that can be modified and get better results if we can properly tune the hyper-parameters.

### 2.4 Efficientnet

In order to categorise handwritten Gujarati, Bangla and Devanagari digits from zero to nine, study [14, 15] demonstrate the effectiveness of transfer learning using EfficientNet.The pre-trained CNN EfficentNet's convolutional and pooling layers, as well as freshly designed fully-connected layers and output layers, were used to create the suggested framework for feature extraction and classification. According to experimental findings in [14] among six pre-trained networks, EfficientNet has obtained the highest accuracy (94.9 percent during training and 94.98 percent during testing). The work [15] also exhibits prominent results in recognizing Bangla Characters that inspires us to employ EfficientNet in this work

## 3 Proposed Methodology

ImageNet contains over 15 million high-resolution human-labeled images belonging to 22,000 classes [16]. As building blocks for our solution, we used three pre-trained ImageNet models, Resnet-50, Inception-v3, and EfficientNetB0. The models had done exceptionally well in the experiments. . We modified some layers of the pre-trained models to adapt to our dataset and objective.

The data samples in our dataset have the dimension of 96 * 96 pixels. The fact that our goal is to recognize the Bengali handwritten numerals suggests that the issue we are attempting to resolve is a categorization issue. For classification problems, the CNN model's output classes must match the dataset's number of classes. So we took all three models mentioned above and then modified their input size to 96*96 and changed the classification layer to have ten output classes in the final layer as we are dealing with ten classification classes.

The following block diagram shows the workflow of our model creation:

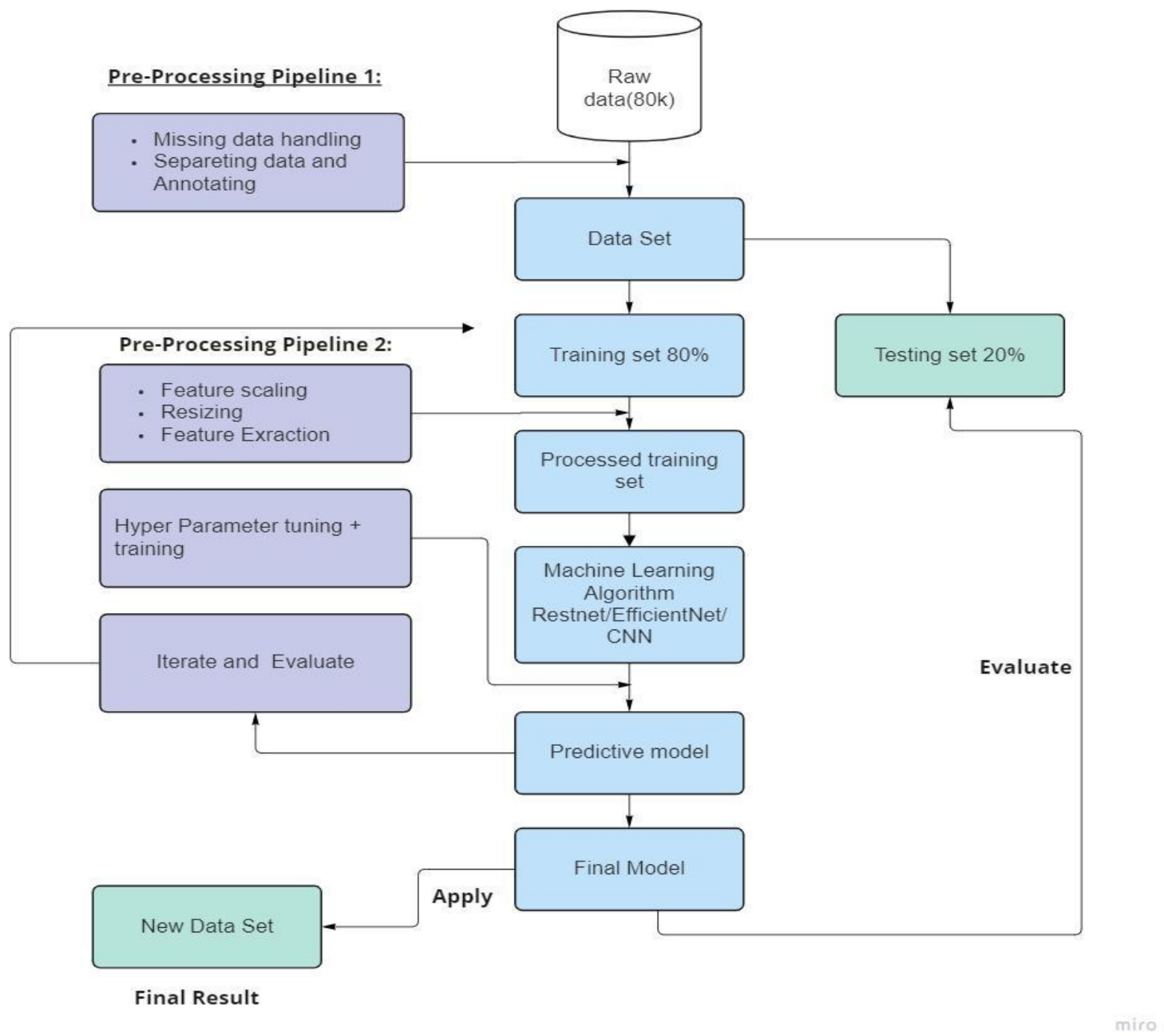

**Figure 1**: Methodological framework

The above diagram shows that we first took the raw version of the dataset. It contained around 80k samples. Then we ran our first pre-processing pipeline, which we identified as pre-processing pipeline1. In this stage, we first handled missing data and then separated the data and annotation. After finishing the pre-processing pipeline1, we created a dataset from the raw dataset, free from any missing data where data and annotations are separated.

Following this, the train-test split was executed on the dataset. Eighty percent of the training set's data and relevant annotations were maintained, while only twenty percent of the testing set's data and annotations were preserved. Our model was trained using the training data. The test set was divided into two sets. We refer to the first as the test set and the second as the new data set. During training, we evaluated our model using the test set; once training was complete, we acquired the result by running the data from a fresh Data set through the trained final model.

As we trained our model on the training set, we applied another pre-processing pipeline to the training set data. In the above diagram, we named it pre-processing pipeline2. We used feature scaling, resizing, and feature extraction in this pipeline to the training data. Then we finally have pre-processed training set. Then finally, we passed this pre-processed data through the machine learning models we have prepared (i.e., ResNet model, EffecientNet model.) Then we evaluated the model using a test set. We iterated this train and evaluation process for many epochs until we achieved sufficient accuracy and flattened loss during the training and evaluating iteration process. Once we had flattened loss and adequate accuracy, we finalized the model. Once we have the finalized model, we passed the New dataset through this finalized model, and we obtained the final result.

## 4 Dataset

We used the NumtaDB dataset [17], which contains more than 85,000 photographs of hand-written Bengali digits, in our suggested methodology. This dataset is a combination of six other datasets, including the Bangla Lekha-Isolated, OngkoDB, DUISRT, Bengali Handwritten Digits Database (BHDDB), and UIUDB. However, every one of the datasets underwent a thorough inspection using the same evaluation standard to ensure that any human being with no prior knowledge could read all the numbers.

The sources are categorized as "a" through "f." Depending on the data source, different subsets of the training and testing sets exist (training-a, testing-a, etc.). All datasets have been split into training and testing sets in order to prevent the inclusion of the handwriting from the same subject or contributor in both sets.

The testing set is supplemented by two enhanced datasets (augmented from test photos of datasets 'a' and 'c'), which include the following augmentations:

1. Rotation, translation, shear, height/width shift, channel shift, and zoom are examples of spatial transformations.
2. Noise, brightness, contrast, saturation, and hue shifts.
3. Obstructions.
4. Superimposition 4. (to create the appearance that text is seen on the opposite side of a page

### 4.1 Pre-processing

Because convolutional neural networks (CNN) are unique in their capabilities, we have avoided heavy pre-processing of images. However, some preprocessing was required for our model to work correctly. We used the built-in **preprocess_input** method available in the Keras library.

Preprocessing a tensor or Numpy array encoding a batch of photos is done using this function. This function also makes the training data compatible with the built-in pre-trained Keras models we intend to use for this paper. How we adopt them will be described in sections.**Pre_process** input supports three input modes for preprocessing. The following figure shows a screenshot of the internal documentation for the process_input method.

```
PREPROCESS_INPUT_MODE_DOC = """
    mode: One of "caffe", "tf" or "torch". Defaults to "caffe".
      - caffe: will convert the images from RGB to BGR,
          then will zero-center each color channel with
          respect to the ImageNet dataset,
          without scaling.
      - tf: will scale pixels between -1 and 1,
          sample-wise.
      - torch: will scale pixels between 0 and 1 and then
          will normalize each channel with respect to the
          ImageNet dataset.
  """
```

**Figure 2:**supported modes for preprocess_input method

The figure above shows that the pre_process input has ‘caffe’, ‘tf’, and ‘torch’ modes. The default mode of the method is ‘caffe’ mode. We used this default mode. This built-in preprocess_input method applied two transformations to the dataset in this mode. First, the images in the dataset were converted to BGR format from RGB format. The goal of this conversion is to prepare the image for OpenCV processing. Then, without scaling, each color channel was zero-centered with respect to the ImageNet dataset. The act of "moving" the distribution's values to make the mean equal to zero is known as zero-centering. The preprocess input method in our particular situation shifts the values of the input picture pixels in relation to the ImageNet dataset without scaling the dataset. As no scaling is used, the samples in the dataset retain the original dimension in pixels.

## 5 Result

This section will discuss the outcomes of each model and then compare them all and finally will recommend the most suitable model for Bangla handwritten digit recognition.

EfficientNet , InceptionV3, and ResNet50 were the three pre-trained models that we implemented. Testing accuracy for EfficientNetBO is 96%, training accuracy is 97%, InceptionV3 testing accuracy is 88%, training accuracy is 90%, RestNet50 testing accuracy is 94%, and training accuracy is 95%. We used 17022k images to get this result.

### 5.1 Evaluation Criteria

For this research work, three CNN models were trained to determine which one performs better in recognizing handwritten Bengali digits. To train the model this research used the Adam optimizer. On an NVIDIA RTX 2080 Ti GPU, the model was trained for 20 experimentally chosen epochs. After training the models, this research utilizes them to predict classes of ten test set images from 0 to 9 in Bangla numeric using those models. To evaluate the results, this research has taken 20 epochs and defined the initial learning rate as 0.0001 for the Inception V3 and EfficientNetB0 model and 1e-4 for the RestNet50 model. Additionally, the batch size was taken at 32 for both RestNet50 and InceptionV3, and 64 for EfficientnetB0. Image height and width were 96 for all three models. To determine the result obtained from these three models this research used the value derived from the classification report which includes the values of precision, recall, and f1 score.

### 5.2 Benchmarking Parameters

1. **Precision**: It indicates how many of the positive predictions are actually positive. It is the percentage of images that are successfully categorized over the total number of images classified.

$$Precision = \frac{TP}{TP+FP}$$

2. **Recal**l: It indicates how many of the actual positives the model accurately predicted. is the percentage of successfully categorized images over all images in class x.

$$Recall = \frac{TP}{TP+FN}$$

3. **F1 score**: It acts as a metric that takes into account both precision and recall, thus when analyzing the results, this research took into account the value of F1, which helped to produce the accuracy, macro avg, and weighted avg results.

$$F1 = 2 \times \frac{Precision \times recall}{Precision + recall}$$

4. **Macro average**: A macro average is a straightforward approach for averaging numeric data that treats all classes in the same way without taking the support value into consideration.
5. **Weighted average**: The weighted average computes the mean of all of the individual f1 values for each class and also takes into account the support value.
6. **Accuracy/Micro average**: The accuracy is determined by taking the global average of the F1-score, which is also known as the micro average, and counts the sum of the true positive, false negative, and false-positive results. Accuracy is another name for the micro average since it computes all of the classes that have been correctly categorized out of the total number of classes.

### 5.3 Result obtained using Inception V3 model:

The classification report is shown below from the test data set using the pre-trained Inception V3 model:

| | precision | recall | f1-score | support |
|---|---|---|---|---|
| 0 | 0.96 | 0.97 | 0.97 | 485 |
| 1 | 0.81 | 0.87 | 0.84 | 477 |
| 2 | 0.90 | 0.91 | 0.90 | 481 |
| 3 | 0.92 | 0.78 | 0.85 | 221 |
| 4 | 0.93 | 0.93 | 0.93 | 402 |
| 5 | 0.86 | 0.90 | 0.88 | 457 |
| 6 | 0.83 | 0.85 | 0.84 | 279 |
| 7 | 0.91 | 0.89 | 0.90 | 201 |
| 8 | 0.94 | 0.91 | 0.93 | 200 |
| 9 | 0.84 | 0.71 | 0.77 | 202 |
| accuracy | | | 0.89 | 3405 |
| macro avg | 0.89 | 0.87 | 0.88 | 3405 |
| weighted avg | 0.89 | 0.89 | 0.89 | 3405 |

**Figure 3:** Inception V3 model Classification report

In the table above, this research considered the value of column F1 asit is crucial to take into account the importance of both precisions and recall in order to comprehend the model's accuracy. This helpedthis research to produce the accuracy, macro avg, and weighted avg results. From the f1 score column, this research obtained a macro value of 0.88, a weighted average and an accuracy of 0.89.As this research chose a dataset where all the classes are equally important, considering the macro average would be good as it considers all the classes without considering any support value.

The classification report provides an introduction to the results, below the line graph is showing the accuracy rate of this model.

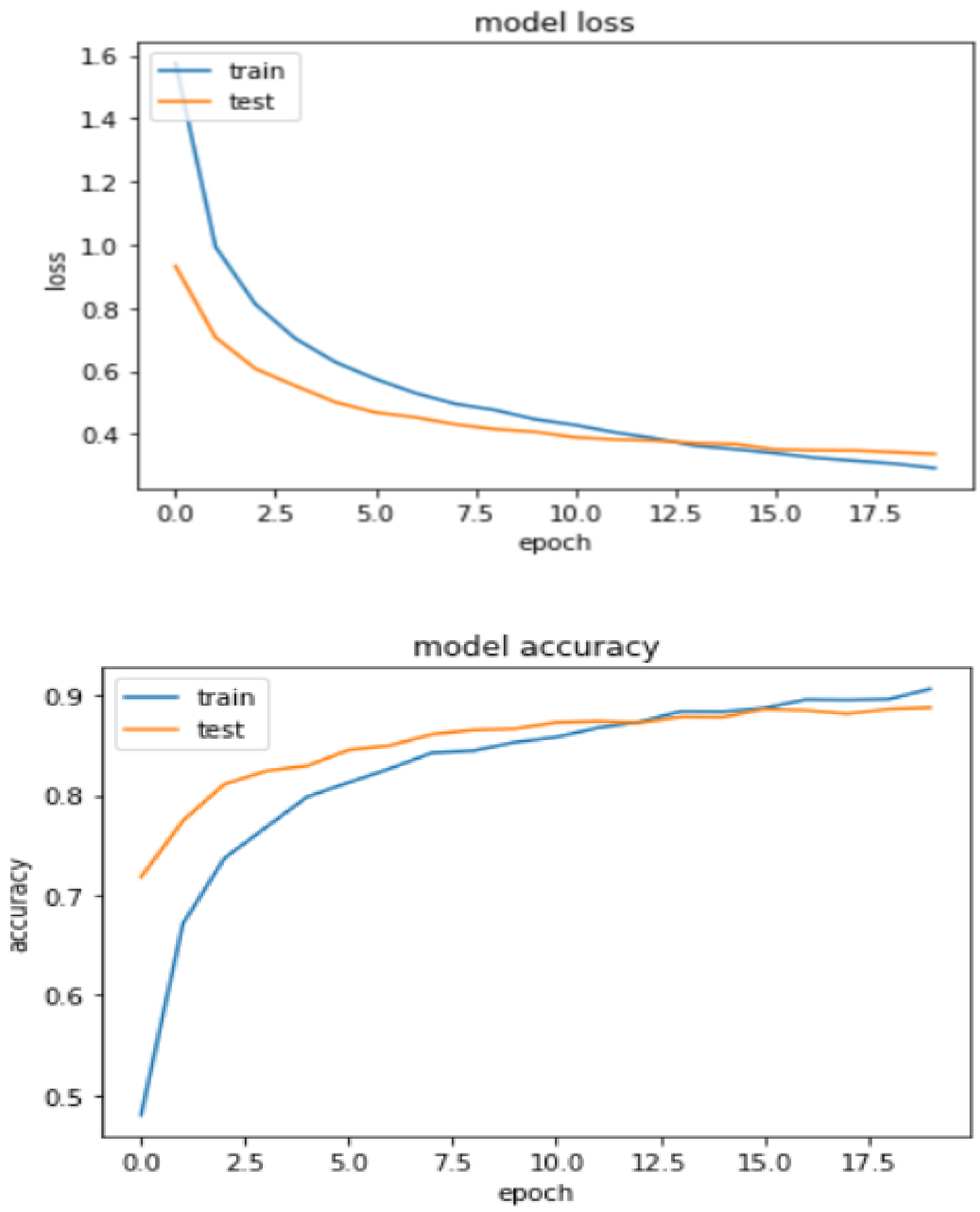

**Figure 4:** Inception V3 model loss and accuracy line graph

From the above line graphs, the results of the train data set are illustrated by the blue line in both images, whereas the results of the test data are shown by the orange line. According to the model loss line chart, at the beginning of our training data set when the epoch was at 1, the loss was higher than usual at 1.6. However, gradually with an increasing amount of epochs, the loss rate reduces, as we can see from the image: for 11 epochs the loss was 0.5, and by the time it reached 20 epochs the loss was approximately 0.1. Also, the initial loss rate in the test data was 0.9, and it gradually fell to 0.4.

In a similar way, the accuracy of the chart model in the training data started off lower, at 0.25, but it rapidly increased all the way up to 0.9 as the number of epochs increased. In addition, the accuracy of the test results began at 0.75 and subsequently grew until it reached around 0.9.

From the above chart, it is observed that test data had less loss and more accuracy rather than the train data which implies after training the data, this model performed well to find the accuracy in test data.

### 5.4 Results obtained from using the EfficientNetB0 model:

The classification report is shown below from the test data set using the pre-trained EfficientB0model:

| | precision | recall | f1-score | support |
|---|---|---|---|---|
| 0 | 0.98 | 0.98 | 0.98 | 485 |
| 1 | 0.97 | 0.95 | 0.96 | 477 |
| 2 | 0.98 | 0.99 | 0.98 | 481 |
| 3 | 0.96 | 0.93 | 0.94 | 221 |
| 4 | 0.99 | 0.98 | 0.98 | 402 |
| 5 | 0.91 | 0.97 | 0.94 | 457 |
| 6 | 0.93 | 0.95 | 0.94 | 279 |
| 7 | 0.98 | 0.98 | 0.98 | 201 |
| 8 | 0.98 | 0.95 | 0.97 | 200 |
| 9 | 0.94 | 0.89 | 0.92 | 202 |
| accuracy | | | 0.96 | 3405 |
| macro avg | 0.96 | 0.96 | 0.96 | 3405 |
| weighted avg | 0.96 | 0.96 | 0.96 | 3405 |

**Figure 5:** EfficientNetB0 model Classification report

EfficientNet-B0 architecture is a mobile-sized architecture having 11M trainable parameters [5]. While considering the EfficientNetB0 model this research took the f1 score into account. As macro avg considers all the classes equally important, thus this research was considering the value of macro avg. However, in this model, all three values are the same 0.96. From these findings, it has been proved that the EfficientNetB0 model is providing the most accuracy in the data set used. The accuracy graph of this model is shown below.

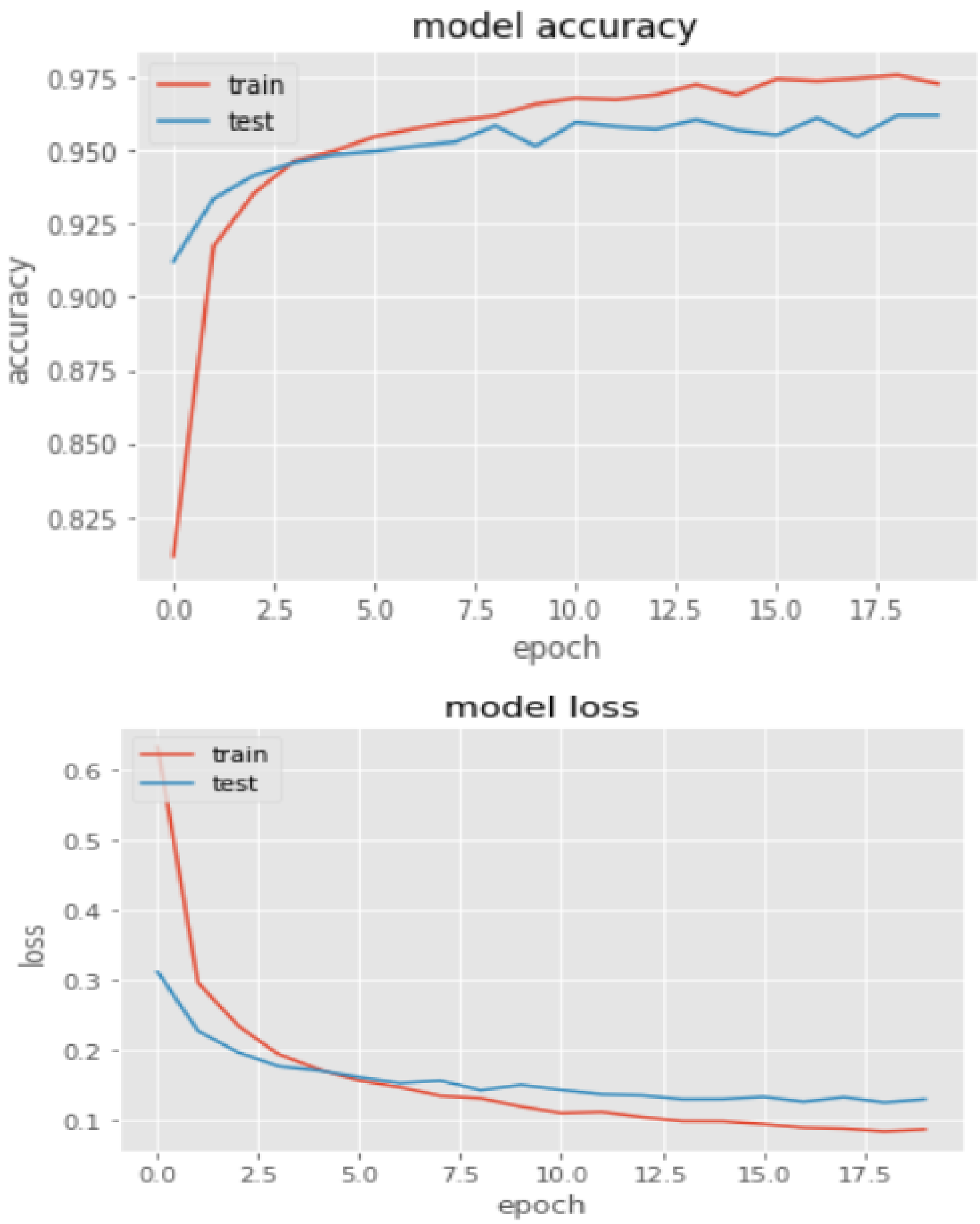

**Figure 6:** EfficientNetB0 model loss and accuracy line graph

In the above line graphs, red lines are showing the train data set and blue lines are showing the test data set. According to the model loss line chart, at the beginning of the training data set when the epoch was at 0, the loss was the highest which was more than0.6 but gradually the loss decreases with increased epochs, and finally, when the epochs reached into 20 the loss was the lowest at less than 0.1. On the other hand, in terms of test data, the model loss was 0.3 at the beginning and it decreased to approximately 0.2 at 20 epochs.

In the accuracy graph, at epoch 1 the accuracy of test data was the lowest at 0.825 but it increased gradually and finally it reached to 0.98 at 20 epochs. In the test data set the accuracy starts from 0.91 at 1 epoch and finally the accuracy reached 0.96 at 20 epochs.

From the above chart, it is observed that test data had less loss and more accuracy in the beginning compared to train data but with an increasing number of epochs the test data loss was more than train data, and accuracy was less than train data. However, overall accuracy in test data was 0.96 which proved that this model present provides a good accuracy rate.

## 5.5 Result obtained using RestNet50 model

Below is the classification report of the test data derived using the pre-trained RestNet50 model.

| | precision | recall | f1-score | support |
|---|---|---|---|---|
| 0 | 0.96 | 0.98 | 0.97 | 485 |
| 1 | 0.92 | 0.91 | 0.92 | 477 |
| 2 | 0.95 | 0.97 | 0.96 | 481 |
| 3 | 0.95 | 0.86 | 0.90 | 221 |
| 4 | 0.96 | 0.97 | 0.96 | 402 |
| 5 | 0.95 | 0.91 | 0.93 | 457 |
| 6 | 0.92 | 0.93 | 0.92 | 279 |
| 7 | 0.92 | 0.98 | 0.95 | 201 |
| 8 | 0.98 | 0.98 | 0.98 | 200 |
| 9 | 0.86 | 0.90 | 0.88 | 202 |
| accuracy | | | 0.94 | 3405 |
| macro avg | 0.94 | 0.94 | 0.94 | 3405 |
| weighted avg | 0.94 | 0.94 | 0.94 | 3405 |

**Figure 7:** RestNet50 Classification report

The ResNet-50 is one of the popular CNN models which have over 23 million trainable parameters. The above classification report considered the f2 score to determine the value of accuracy, macro avg, and weighted avg, and this pre-trained model produced the same value of 0.94 for all three components.

The loss and accuracy rate of this model is shown in the line graph below.

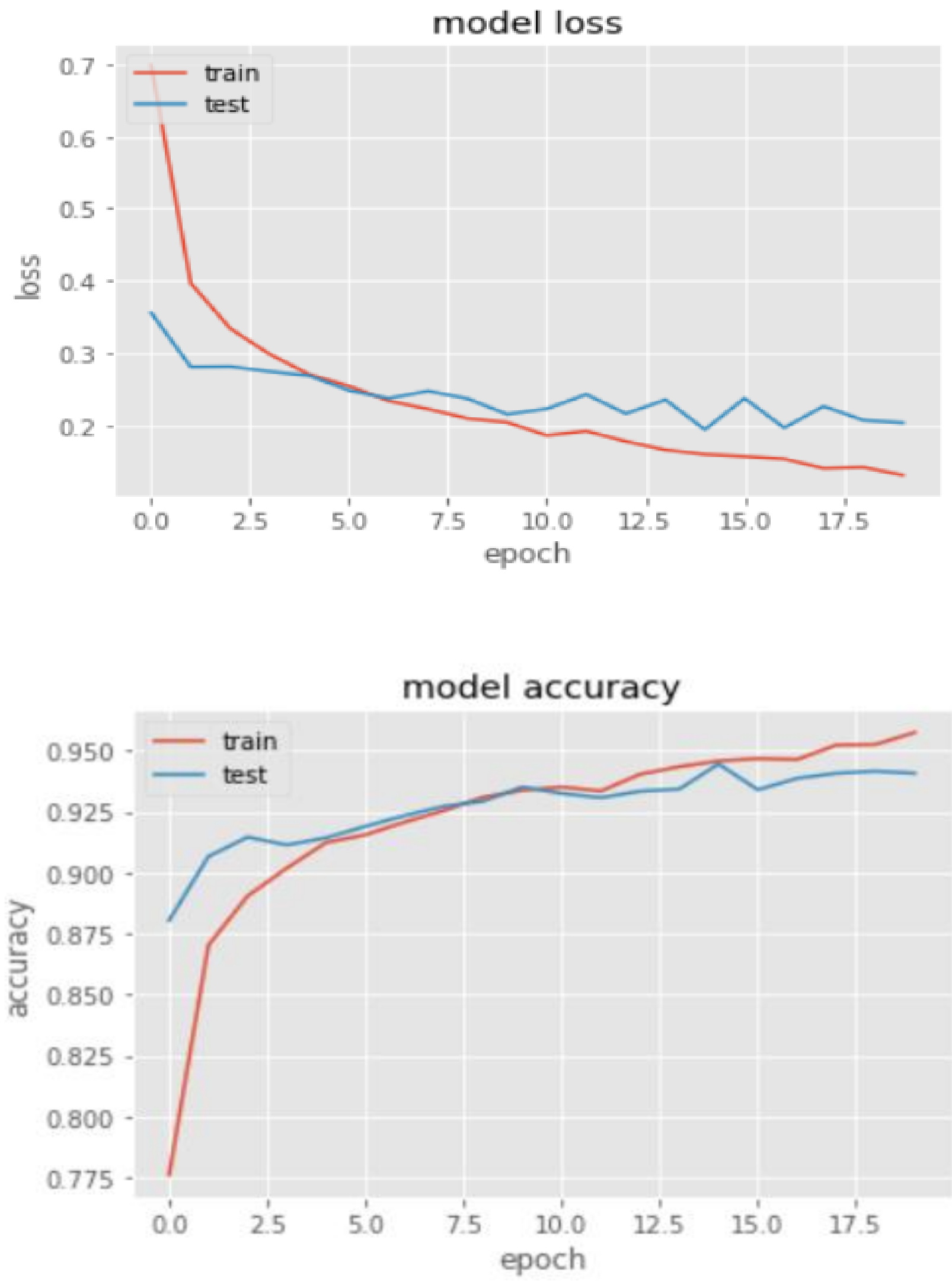

**Figure 8:** RestNet50 model loss and accuracy line graph

In the above graphs, Train results are represented by the red line, and test results are presented by the blue line. According to the loss graph, the loss was 0.7 at the start of the training data set at epoch1and it gradually decreased to 0.1 when it reached 20 epochs. In terms of test data, at 1 epoch loss was more than 0.3 and at 20 epoch it reached 0.2. In between the lines were more curved where the loss fluctuated.

In accordance with the accuracy graph, at 1 epoch the accuracy was the lowest for the train data set at 0.78 which increased to 0.96 at 20 epochs. However, the initial accuracy for test data starts from 0.88 at 1 epoch and reached up to 0.94 at 20 epochs.

The summary of those two plots demonstrated that the model behaved better following training, resulting in relatively stable initial loss and accuracy at end epochs.

## 6 Discussion

In this research, only one data set was used to experiment with three of the pre-trained models, to understand the efficiency of their performance. After collecting the test results from each of the models, this study considered the macro average value since it takes into account all of the classes if the data set is not unbalanced. This study additionally evaluates the accuracy value, as the total number of true positive, false negative, and false-positive cases are tallied, and finally, it also reflects theweightedd avg value as it focused on the proportion of each class's support value while comparing with the sum of all support values.

In order to determine which model is the most appropriate and has the highest level of accuracy, it was crucial for this research to compare all three components of calculating the accuracy.The below graph shows the macro avg, accuracy, and weighted avg comparison of all three models.

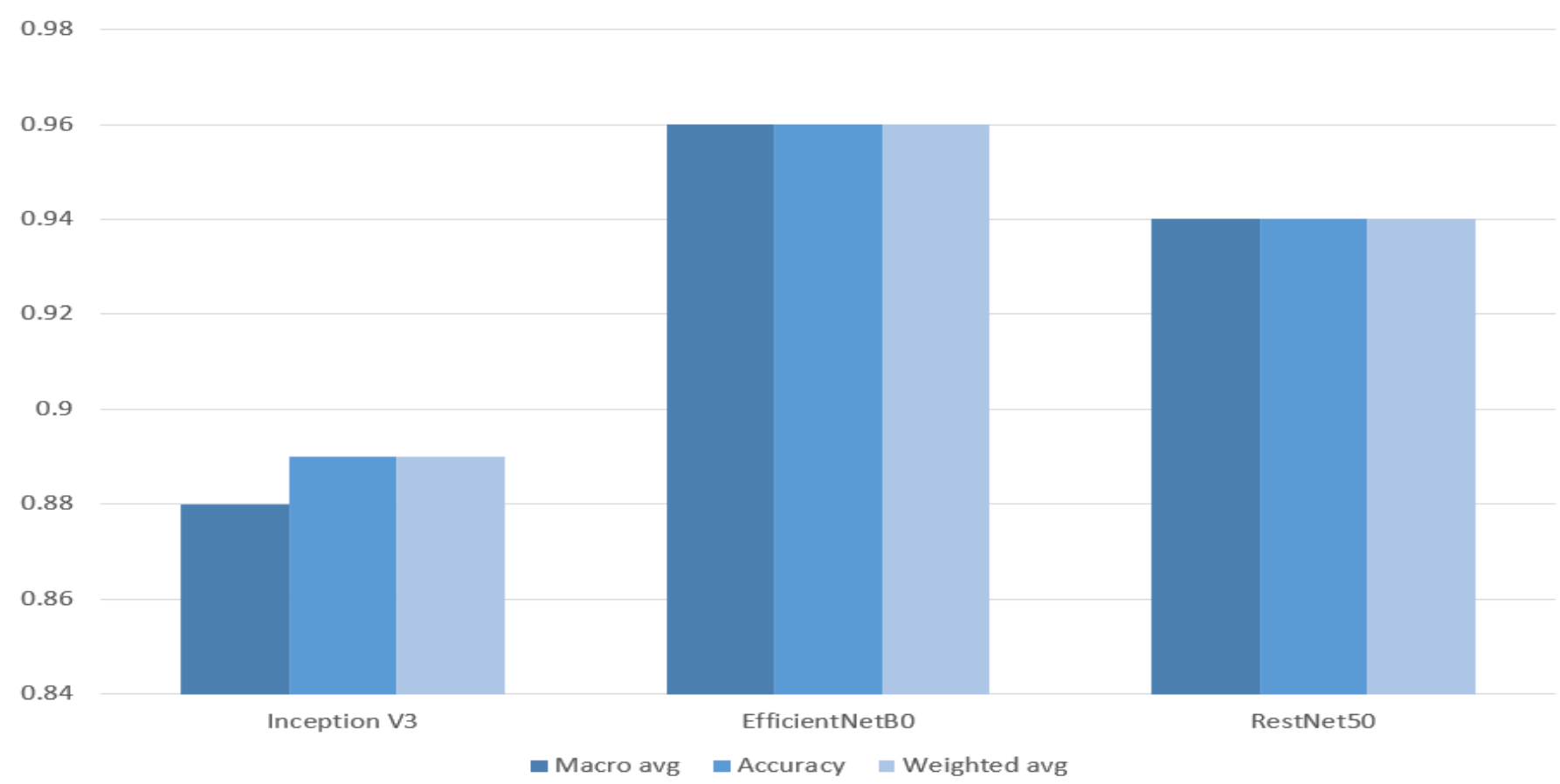

**Figure 9:** Comparison of all three models against the benchmark derived from the classification report

The above graph explained that Both EfficientNetB0 and RestNet50 have a higher accuracy rate than the Inception V3 model. EfficientNetB0 has the highest accuracy 0.96 compared to all three models. So, it can be said that this pre-trained model worked the best rather than the other two models. The below line graph is showing the comparison of accuracy and loss within three pre-trained models within test data when it reached 20 epochs.

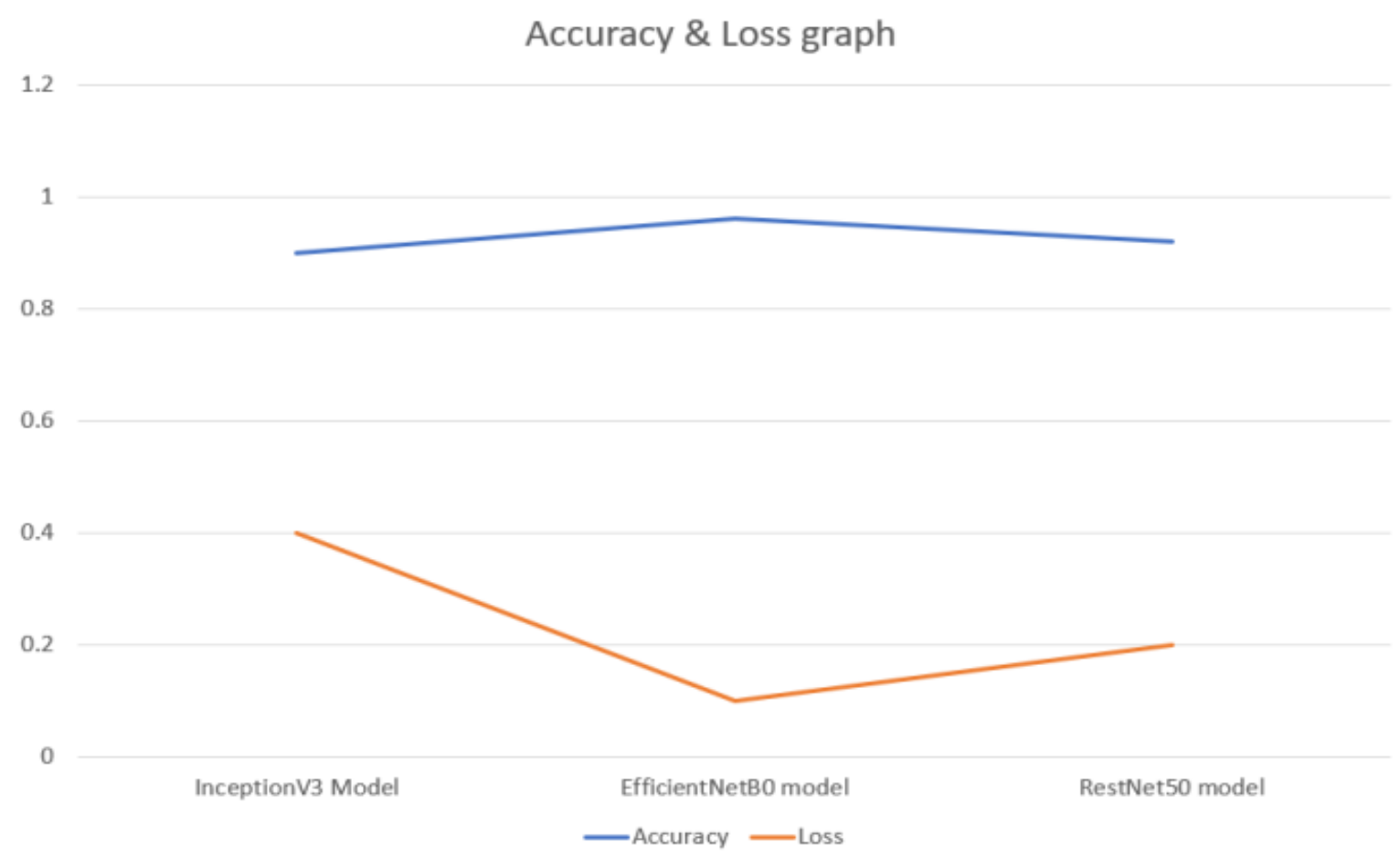

**Figure 10:** Comparison of accuracy and loss between three pretrained model

This research has found that the InceptionV3 model had the highest loss when it reached 20 epochs with the test data and EfficientNetB0 had the lowest loss at 0.1. Moreover, EfficientNetB0 had the highest accuracy at 0.96 whereas RestNet50 had 0.91. In summary, it can be said InceptionV3 model had the highest loss and lowest accuracy, whereas the RestNet50 model had quite similar loss and accuracy. Efficient Net provided the highest accuracy and lowest loss, as a result, this model performed best while recognizing the Bangla digit.

From the results of three pre-trained models, this study also found that the loss rate was lower, and the accuracy rate was higher in test data, after training the train data set. Moreover, the difference between the loss from 1 epoch to 20 epochs was higher in the training data set but this difference reduces after training the data and test data showed fewer differences across the epochs. The same results were found in the accuracy graph as well. From this, it can be mentioned that if the models can be trained properly, the models will provide more accurate results in test data.

Considering the accuracy graph of the three models, the difference between 1 epoch to 20 epochs within test data is shown in the below table.

| Model | Accuracy at 1 epoch | Accuracy at 20 epochs | Difference |
|---|---|---|---|
| **Inception V3** | 0.75 | 0.90 | 0.15 |
| **EfficientNetB0** | 0.91 | 0.96 | 0.05 |
| **RestNet50** | 0.88 | 0.94 | 0.06 |

**Figure 11:**Difference in accuracy considering the test data from 1 epoch to 20 epochs for all three models.

The above table represents that EfficientNetB0 has the capacity to act better on training and is able to provide more accuracy compared to the other two models. It also proves that with proper training this model performs best recognition and the accuracy for all the epochs within a test data set is close to each other, which means it provided almost the same accuracy while recognizing the Bangla digit from 0 to 9.

According to the results and above discussion this research suggests EfficientNetB0 as the best CNN pre-trained model to use to recognize Bangla digits.

## 7 Contributıon

The main contribution of our research includes-

1. Proposing the most efficient pre-trained CNN model for recognizing Bengali digits.
2. Identifying the pre-trained model which is able to provide the highest accuracy while recognizing different Bengali digits images.
3. Identifying which pre-trained model can be trained more efficiently

## 8 Conclusion

The Bengali language is the seventh most spoken language in the world, yet little research has been done on it. It is also feasible to write Bengali numbers differently due to their intricate shapes. Thus, it is essential to integrate a model of image recognition capable of identifying the Bengali digit. This research article aimed to develop a deep learning model capable of identifying hand-drawn representations of Bengali numerals. This research study demonstrates that the Convolutional Neural Network successfully recognized hand-drawn Bengali numerals. In addition, the purpose of this study was to determine which model inside the CNN model is more successful and delivers greater accuracy when recognizing Bengali digits. After testing three pre-trained models, Inception V3, EfficientNetB0, and RestNet50, this research concludes that the EffficientNetB0 model delivers the best outcome. This model could be trained more effectively than the other two, providing greater accuracy and less loss, and the accuracy of each component of the data set does not vary significantly, resulting in a higher accuracy rate for all Bengali digits.

## 9 FUTURE WORK

This research implemented a single data set and three pre-trained CNN models to recognize Bangla digits and determine which model offers the maximum accuracy. In the future, this research intends to examine the performance of these pre-trained models when working with several other datasets. In addition, it intends to evaluate the precision of all pre-trained CNN models not specified in this study, such as VGG19. These future studies will increase the opportunities for future researchers and academics to utilize CNN models.

**Author Contributions** Islam M, significantly contributed to the conceptual parts of the paper's contribution to the knowledge. Then Islam M. and Shuvo SA devised the methodology for the paper and conducted the experiment. Shaikh MM and Sourav MSU contributed to the section on methodology. Nipun MS and Nayeem J have examined the outcomes of the experiment and contributed to the section's design and writing. Abstract, introduction, and conclusion were all written by Haque Z. Sulaiman RB oversaw the entire project and contributed to every aspect of this study article. Shuvo SA has contributed to the administration of group collaboration on a global scale. Sourav MSU assisted in finalising the manuscript. All authors examined the findings and approved the final manuscript version.